\documentclass{article}

\usepackage{enumitem}

\usepackage[preprint,nonatbib]{neurips_2019}
\usepackage{authblk}
\usepackage[utf8]{inputenc} 
\usepackage[T1]{fontenc}    
\usepackage{hyperref}       
\usepackage{url}            
\usepackage{booktabs}       
\usepackage{amsfonts}       
\usepackage{nicefrac}       
\usepackage{microtype}      
\usepackage{eso-pic}
\usepackage{xspace}
\usepackage{amsmath}
\usepackage[utf8]{inputenc}
\usepackage[english]{babel}
\usepackage{appendix}
\usepackage{graphicx}
\usepackage{amssymb}
\usepackage{bm}
 
\newtheorem{theorem}{Theorem}[section]
\newtheorem{corollary}{Corollary}[theorem]
\newtheorem{lemma}[theorem]{Lemma}

\makeatletter
\DeclareRobustCommand\onedot{\futurelet\@let@token\@onedot}
\def\@onedot{\ifx\@let@token.\else.\null\fi\xspace}

\def\ie{\emph{i.e}\onedot} 
 
\def\etc{\emph{etc}\onedot} 
 
\def\etal{\emph{et al}\onedot}
\makeatother

\title{Semi-supervised Learning using Adversarial Training with Good and Bad Samples}

\author[1]{Wenyuan Li}
\author[1]{Zichen Wang}
\author[2]{Yuguang Yue}
\author[1]{Jiayun Li}
\author[1]{William Speier}
\author[2]{Mingyuan Zhou}
\author[1]{Corey W. Arnold}

\affil[1]{University of California, Los Angeles}
\affil[2]{University of Texas, Austin}

\begin{document}
\maketitle
\vspace{-3mm}\begin{abstract}\vspace{-3mm}
In this work, we investigate semi-supervised learning (SSL) for image classification using adversarial training. Previous results have illustrated that generative adversarial networks (GANs) can be used for multiple purposes. Triple-GAN, which aims to jointly optimize model components by incorporating three players, generates suitable image-label pairs to compensate for the lack of labeled data in SSL with improved benchmark performance. Conversely, Bad (or complementary) GAN, optimizes generation to produce complementary data-label pairs and force a classifier's decision boundary to lie between data manifolds. Although it generally outperforms Triple-GAN, Bad GAN is highly sensitive to the amount of labeled data used for training. Unifying these two approaches, we present unified-GAN (UGAN), a novel framework that enables a classifier to simultaneously learn from both good and bad samples through adversarial training. We perform extensive experiments on various datasets and demonstrate that UGAN: 1) achieves state-of-the-art performance among other deep generative models, and 2) is robust to variations in the amount of labeled data used for training.
\end{abstract}

\vspace{-3mm}\section{Introduction}\vspace{-3mm}
With recent progress in deep learning, large labeled training datasets are becoming increasingly important \cite{deng2009imagenet, lin2014microsoft, abu2016youtube, krasin2017openimages}. However, labeling such datasets is expensive and time-consuming. Semi-supervised learning (SSL) aims to leverage large amounts of unlabeled data to boost model performance. Various SSL methods have been proposed using deep learning and proven to be successful. Weston \etal \cite{weston2012deep} employed a manifold embedding technique using a pre-constructed graph of unlabeled data; Rasmus \etal \cite{rasmus2015semi} used a specially designed auto-encoder to extract essential features for classification; Kingma and Welling \cite{kingma2013auto} developed a variational auto encoder by maximizing the variational lower bound of both labeled and unlabeled data; Miyato \etal \cite{miyato2018virtual} proposed virtual adversarial training (VAT), which helped find a deep classifier that had a good prediction accuracy and was less sensitive to data perturbation towards the adversarial direction.

Recently, generative adversarial networks (GANs) \cite{goodfellow2014generative}, have demonstrated their capability in SSL frameworks \cite{salimans2016improved, dai2017good, gan2017triangle, chongxuan2017triple, kumar2017semi, lecouat2018semi, li2018semi}. GANs are a powerful class of deep generative models that can represent data distributions over natural images \cite{radford2015unsupervised, mirza2014conditional}. Specifically, a GAN is formulated as a two-player game, where the generator $G$ takes a random vector $z$ as input and produces a sample $G(z)$ in the data space, while the discriminator $D$ identifies whether a certain sample comes from the true data distribution $p(x)$ or the generator. 
As an extension, Salimans \etal \cite{salimans2016improved} first proposed feature-matching GANs (FM-GANs) to solve an SSL problem. Suppose we have a classification problem that requires classifying a data point $x$ into one of $K$ possible classes. A standard classifier takes $x$ as input and outputs a $K$-dimensional vector of logits $\{l_1, ..., l_K\}$. Salimans \etal extended the standard classifier by simply adding samples from a GAN's $G$ to the dataset, labeling them as a new ``generated'' class $y = K + 1$, and correspondingly increasing the  classifier's output dimension  from $K$ to $K + 1$. They also found that using feature matching loss in $G$ improved classification performance. The $(K + 1)$-class discrimination objective with feature matching loss in $G$ led to strong empirical results.

Empirically, FM-GANs demonstrate good performance on SSL classification tasks; however, the generated images from the generator are low-quality, \ie, the generator may create visually unrealistic images. Li \etal \cite{chongxuan2017triple} realized that the generator and the discriminator in FM-GANs may not be optimal at the same time. Intuitively, assuming the generator can create good samples, the discriminator should identify these samples as fake samples as well as predict the correct class for them. To address this problem, they proposed a three-player game, Triple-GAN, to simultaneously achieve superior classification results and obtain a good image generator. Triple-GAN consisted of a generator \textit{G}, a discriminator \textit{D}, and a separate classifier \textit{C}. \textit{C} and \textit{G} were two conditional networks that generated pseudo labels given real data, and pseudo data given real labels, respectively. To jointly evaluate the quality of the samples from the two conditional networks, \textit{D} was used to distinguish whether a data-label pair was from the real labeled dataset or not. The improvements achieved by Triple-GAN were more significant as the number of labeled data decreased, suggesting that the generated data-label pairs can be used effectively to train the classifier. Meanwhile, Dai \etal \cite{dai2017good} realized the same problem of the generator, but instead gave theoretical justifications of why using ``bad'' samples from the generator could boost SSL performance. Loosely speaking, they defined samples that form a complement set of the true data distribution in feature space as ``bad'' samples. By carefully defining the generator loss, the generator could create ``bad'' samples that forced \textit{C}'s decision boundary to lie between the data manifolds of different classes, which in turn improved generalization of $C$. Their model was called Bad GAN, which achieved state-of-the-art performance on multiple benchmark datasets. Most recently, Li \etal \cite{li2019semi} performed a comprehensive comparison between Triple-GAN and Bad GAN. They illustrated the distinct characteristics of the images the models generated, as well as each model’s sensitivity to various amount of labeled data used for training. Furthermore, they showed that in the case of low amounts of labeled data, Bad GAN's performance decreased faster than Triple-GAN, and both models' performance were contingent on the selection of labeled samples; in other words, selecting non-representative samples would deteriorate the classification performance.

In this paper, we present unified-GAN (UGAN), a semi-supervised learning framework that unifies both good and bad generated samples and takes advantage of them through adversarial training. Inspired by Triple-GAN and Bad GAN, we find that good and bad synthetic samples can be used for complementary purposes. Generated good image-label pairs can be used to train the classifier, while the bad samples can force the decision boundary to be between the data manifold of different classes. Hence, we leverage both good and bad generated samples in the proposed UGAN and achieve further performance improvement in SSL. Overall, our main contributions of this paper are: 
1) we propose a novel SSL framework, UGAN, which simultaneously trains a good and bad generators through adversarial training and takes advantage of both generated samples to boost SSL performance;
2) we analyze our proposed UGAN, theoretically prove its global optimum, and additionally put UGAN in the Expectation-Maximization (EM) framework and validate its non-increasing divergence property; and
3) we do extensive experiments to show that UGAN can improve upon state-of-the-art classification results in SSL, and show the effectiveness of the model with different amounts of labeled data.

\begin{figure*}
\begin{center}
  \includegraphics[width=13.5cm]{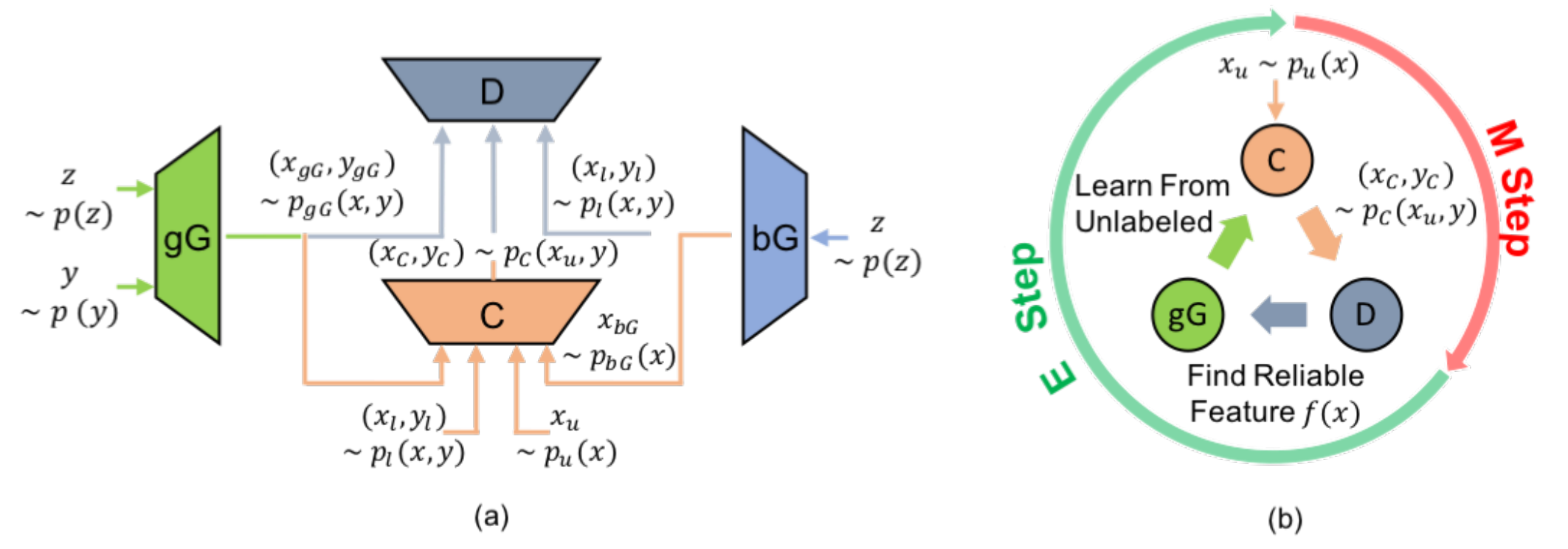}\vspace{-3mm}
  \caption{(a) Network architecture of UGAN. UGAN consists of four components: 1) a bad generator, $bG$, generates ``bad'' samples; 2) two conditional networks, $gG$ and $C$, that generate pseudo labels given real data, and pseudo data given real labels, respectively; and 3) a separate discriminator, $D$, that distinguishes the generated data-label pair from the real data-label pair. (b) EM analysis of UGAN. During the E step, $C$ predicts labels for unlabelled data, and then randomly selects some unlabeled data and uses these pairs as positive samples for $D$; by doing so, $gG$ is able to generate pseudo pairs $(x_{gG},y_{gG})$ that implicitly contain features from unlabeled data. During the M step, $(x_{gG},y_{gG})$ are used to minimize $\text{KL}(p(y|x)||p_\theta(y|x,y\leq K))$.} \label{Figure1}\vspace{-9mm}
\end{center}
\end{figure*}

\vspace{-4mm}\section{Related Work}\vspace{-4mm}
Besides the aforementioned FM-GAN \cite{salimans2016improved}, Triple-GAN \cite{chongxuan2017triple}, and Bad GAN \cite{dai2017good}, several previous studies have also incorporated the idea of adversarial training in SSL. CatGAN \cite{springenberg2015unsupervised} substituted the binary discriminator in standard GAN with a multi-class classifier and trained both the generator and discriminator using information theoretical criteria on unlabeled data. Virtual adversarial training (VAT) \cite{miyato2018virtual} effectively smoothed the classifier output distribution by seeking virtual adversarial samples. In adversarial learned inference \cite{dumoulin2016adversarially}, the inference network approximated the posterior of latent variables given true data in an unsupervised manner. Another line of work has focused on manifold regularization \cite{belkin2006manifold}. Kumar \etal \cite{kumar2017semi} estimated the manifold gradients at input data points and added an additional regularization term to a GAN, which promoted invariance of the discriminator to all directions in the data space. Lecouat \etal \cite{lecouat2018semi} achieved competitive results by performing manifold regularization using approximate Laplacian norm that was easily computed within a GAN.

Apart from adversarial training, there have been other efforts in SSL recently. $\Gamma$ model  \cite{rasmus2015semi} evaluated unlabelled data with and without noise, and applied a consistency cost between the two predictions. It assumed a dual role as a teacher and a student. The teacher generated targets of unlabeled data, which were then used to train a student. Since the model itself generated the targets, they could be incorrect. To alleviate the problem, $\Pi$ model \cite{laine2016temporal} added noise at the inference time, and consequently a noisy teacher could yield more accurate targets. $\Pi$ model was further improved by Temporal Ensembling \cite{laine2016temporal}, which maintained an exponential moving average (EMA) prediction for each of the training examples. Consequently, the EMA prediction of each example was formed by an ensemble of the model’s current version and those earlier versions that evaluated the same example. This ensembling improved the quality of the predictions, and using the predictions as teacher signals improved results. Mean Teacher \cite{tarvainen2017mean} averaged model weights to form a target-generating teacher model. Unlike Temporal Ensembling, Mean Teacher worked with large datasets and on-line learning, which was able to improve the speed of learning and classification accuracy simultaneously.

Our proposed UGAN is mainly inspired by Triple-GAN and Bad GAN, these models can be used for complementary purposes. Nevertheless, it has a connection with those ``teacher'' models, as will be seen in Section \ref{Methods}, our model provides a smart way to generate input-label pairs and use them as teaching signals to improve the SSL results.

\vspace{-3.5mm}\section{Method}\label{Methods}\vspace{-3.5mm}
To outline our approach, we consider the same SSL problem as in Triple-GAN \cite{chongxuan2017triple} and Bad GAN \cite{dai2017good}. Given a relatively small labeled set $(x_l, y_l) \sim p_l(x, y) $, where $y \in \{1, 2, \cdot \cdot \cdot, K\}$ is the label space for classification, and a large unlabeled set $x_u \sim p_u(x)$, the goal is to utilize the large amount of unlabeled data to predict the labels $y$ of the unseen samples. 
Suppose the true data distribution is denoted as $p(x, y)$,  we aim to obtain a classifier that can approximate the conditional distribution $p_{C}(y|x) \approx p(y|x)$. To achieve this, we will use an adversarial training process that enables the classifier to learn from both good and bad samples. Specifically, a good generator is able to generate good image-label pairs to train the classifier, while a bad generator generates samples that force the classifier's decision boundary between the data manifolds of different classes. As will be shown, our model takes advantage of both good and bad synthetic samples, and inherits the good properties of both Triple-GAN and Bad GAN. 

\vspace{-2mm}\subsection{Adversarial Training Process with Four Players}\vspace{-2mm}
Our model consists of four parts: 1) a good generator, $gG$, that characterizes the conditional distribution $p_{gG}(x|y) \approx p(x|y)$; 2) a bad generator, $bG$, that takes in a latent vector $z$ and outputs ``bad'' samples \cite{dai2017good}; 3) a classifier, $C$, that characterizes the conditional distribution $p_c(y|x) \approx p(y|x)$; and 4) a discriminator, $D$, that distinguishes whether a pair of data $(x, y)$ comes from the true distribution $p(x, y)$ or not. All the components are parameterized as neural networks, as shown in Fig.~\ref{Figure1}~(a).

We follow  Li \etal \cite{chongxuan2017triple} and assume that the samples from both real data $p(x)$ and real label $p(y)$ can be easily obtained.\footnote{In semi-supervised learning, $p(x)$ is the empirical distribution of inputs and $p(y)$ is assumed same to the distribution of labels on labeled data, which is uniform in our experiments.} In our model, $gG$ produces a pseudo input-label pair by first drawing $y \sim p(y)$ and latent vector $z \sim p(z)$ (we use a uniform distribution for $z$ in our experiments), and then generating $x_{gG} \sim p_{gG}(x|y,z)$. $bG$ generates bad samples by transforming the latent vector $z \sim p(z)$ as in a traditional GAN to obtain $x_{bG} \sim p_{bG}(x|z)$. $C$ takes in four different types of samples (\ie, labeled data, unlabeled data, samples from $gG$, and samples from $bG$) and produces pseudo labels $y$ for them following the conditional distribution $p_{C}(y|x)$. For the labeled data $x_l$, and the $gG$ generated samples $x_{gG}$, we anticipate $C$ to put them into the right class (\ie, either the class $y_l$ of the labeled data $x_l$, or the conditional labels $y$ based on which $x_{gG}$ are generated). For the generated samples from $bG$ $x_{bG}\sim p_{bG}(x|z)$, and unlabeled data $x_u \sim p_u(x)$, we anticipate $C$ to put them into the $(K+1)$th class (\ie the ``fake'' class) and one of the $K$ classes of real data, respectively. Due to the fact that the softmax layer is over-parameterized, we can still model $C$ with $K$ neurons at the output layer by modifying the loss function (see details in Appendix \ref{app_softmax}). $D$ accepts the input-label pairs generated by both $C$ $(x_{C}, y_{C}) \sim p(x_u)p_C(y|x_u)$, and $gG$ $(x_{gG}, y_{gG}) \sim p(y)p_{gG}(x|y)$, and the pairs from the labeled data distribution $(x_l, y_l) \sim p_l(x, y)$ for judgement. $D$ treats the labeled data pairs as  positive samples, while the pairs from both $gG$ and $C$ as negative.  We refer the loss function of $gG$ as\footnote{In practice, we use $L_{gG} = -\mathbb{E}_{x, y\sim p_{gG}(x, y)}[\log (p_{D}(x, y)]$ to ease the training process \cite{goodfellow2014generative}.}
\begin{equation} \label{good_generator_loss}
\begin{aligned}
L_{gG} &= \mathbb{E}_{x, y\sim p_{gG}(x, y)}[\log (1 - p_{D}(x, y)]\\
\end{aligned}
\end{equation}

The loss function of $bG$ is 
\begin{equation} \label{bad_generator_loss}
\begin{aligned}
L_{\textit{bG}} &= -\mathcal{H}(p_{bG}(x)) + \left\| \mathbb{E}_{x\sim p_{\textit{u}}(x)}(\boldsymbol{f}(x)) - \mathbb{E}_{x \sim p_{bG}(x)}(\boldsymbol{f}(x)) \right\|_2^2 \\
\end{aligned}
\end{equation}
where $ -\mathcal{H}(p_{bG}(x))$, which measures the negative entropy of $bG$ generated samples, is used to avoid collapsing while increasing the coverage of $bG$. The second term is feature matching loss, where $\boldsymbol{f}(x)$ denotes a feature map of an intermediate layer of $C$. $D$'s loss function becomes
\begin{equation} \label{discriminator_loss}
\begin{aligned}
L_{D} = & -\mathbb{E}_{x, y \sim p_{l}(x, y)}[\log (p_{D}(x, y)]  - \frac{1}{2}\mathbb{E}_{x, y \sim p_{gG}(x, y)}[\log (1 - p_{D}(x, y)]\\
&  - \frac{1}{2} \mathbb{E}_{x, y\sim p_{C}(x, y \leq K)}[\log (1 - p_{D}(x, y)]\\
\end{aligned}
\end{equation}
where $D$ treats the labeled data as positive samples, and the pseudo input-label pairs from both $gG$ and $C$ as negative samples. Finally, the loss function of $C$ consists of four components,
\begin{equation} \label{classifier_loss_sep}
\begin{aligned}
L_{C_{1}} &= -\mathbb{E}_{x, y\sim p_{l}(x, y)}[\log (p_{C}(y|x, y \leq K)]
& L_{C_2} &=-\mathbb{E}_{x, y\sim p_{gG}(x, y)}[\log (p_{C}(y|x, y \leq K)]\\
L_{C_{3}} &= -\mathbb{E}_{x \sim p_{u}(x)}[\log (1 - p_{C}(y = K + 1 | x)]
& L_{C_{4}} &= -\mathbb{E}_{x \sim p_{bG}(x)}[\log ( p_{C}(y = K + 1 | x)]\\
\end{aligned}
\end{equation}
and the total loss for $C$ is 
\begin{equation} \label{classifier_loss}
L_{C}  = L_{C_1} + \lambda_0 L_{C_2} + \lambda_1 L_{C_3} + \lambda_2 L_{C_4}
\end{equation}
where $L_{C_1}$ and $L_{C_2}$ denote the cross entropy loss for labeled and $gG$ generated samples, respectively, $L_{C_3}$ forces $C$ to put the unlabeled data into real classes, while $L_{C_4}$ forces $C$ to put the $bG$ generated samples into the ``fake'' class. $\lambda_{0, 1, 2}$ is a hyperparameter used to balance each loss component.

The model defined by (\ref{good_generator_loss})-(\ref{classifier_loss}) achieves its equilibrium if and only if $p(x, y) = p_{gG}(x, y) = p_{C}(x, y \leq K)$. In other words, incorporating the bad samples does not change the equilibrium point of Triple-GAN (see Section \ref{global_optimum}). Our model consists of three adversarial parts: 1) $gG$ tries to fool $D$ by generating realistic images conditioned on label $y$; 2) $C$ tries to fool $D$ by generating good labels for unlabeled images; and 3) $bG$ tries to fool $C$ by generating images that are close to the data manifold. At convergence, $D$ cannot distinguish both $ p_{gG}(x, y)$ and $p_{C}(x, y)$ from the true data distribution $p(x, y)$, which indicates that we have obtained both a good $gG$ and a good $C$. Bad samples from $bG$ accelerate this process and improve the generalization of $C$. 

One key problem of SSL is the limited amount of labeled data. A powerful $D$ may memorize the empirical distribution of the labeled data, and reject other types of samples from the true data distribution. Limited labeled data also restricts $gG$ to explore a larger space of the true data distribution. To address this problem, we adopt the practical techniques in Li \etal \cite{chongxuan2017triple}. We generate pseudo labels through $C$ for some unlabeled data and use these pairs as positive samples of $D$. This introduces some bias to the target distribution of $D$, but using the EM framework to analyze the training procedure (see Section \ref{EM_framework}), we are able to prove the rationality of this choice. Moreover, since $C$ converges quickly, this operation provides a way to enable $gG$ to explore a much larger data manifold that includes both the labeled and unlabeled data information. As illustrated in Fig. \ref{Figure1} (b), $C$ is able to provide pseudo labels for the unlabeled data, while $D$ will judge if the pseudo labels are reliable or not. This in return will affect the evolution of $gG$ that will take advantage of the unlabeled data to generate good images. Generated good image-label pairs that implicitly contain unlabeled data information will eventually benefit $C$. This works extremely well for relatively simple datasets like MNIST, and under the circumstance where only an extremely low amount of labeled data is available.

\vspace{-2mm}\subsection{Theoretical Analysis}\vspace{-2mm}
We now give theoretical justification for our four-player game based on the loss functions as mentioned above. We mainly focus on two important properties of our model: 1) the global optimum of the game is the true distribution, which satisfies $p(x,y) = p_{gG}(x,y) = p_C(x,y|y\leq K)$; and 2) the KL divergence between the conditional density of $C$ and the true density, KL$(p(y|x) || p_C(y|x,y\leq K))$, is non-increasing after each iteration when we assume the maximum likelihood estimate (MLE) of $C$ is obtained. A detailed proof of these properties is provided in Appendix \ref{app_theoretical_analysis}.
\vspace{-1mm}\subsubsection{Global Optimum} \label{global_optimum}\vspace{-1mm}
We first show that the optimal $D$ balances between the true data distribution and the mixture distribution defined by $C$ and $gG$, as summarized in Lemma \ref{optD}.
\begin{lemma}\label{optD}
For any fixed $C$ and $gG$, the optimal $D$ of the game defined by loss functions (\ref{good_generator_loss})-(\ref{classifier_loss_sep}) is
\begin{equation}
    D^*_{C,gG,bG}(x,y) = \frac{p_l(x,y)}{p_l(x,y)+p_{\frac{1}{2}}(x,y)},
\end{equation}
where $p_{\frac{1}{2}}(x,y) = \frac{1}{2}p_{gG}(x,y) + \frac{1}{2}p_{C}(x,y|y\leq K)$.
\end{lemma}
Given $D^*_{C,gG,bG}$, we can plug in the optimal $D^*$ in (\ref{discriminator_loss}) and get a value function $V(C,gG,bG)$. Then we have:

\begin{theorem}
The global minimum of $V(C, gG, bG)$ is achieved only when $p_l(x,y)=p_{gG}(x,y)=p_C(x,y|y\leq K)$. 
\end{theorem}

We now consider the case for $p_C(y=K+1|x)$ with the following Corollary \ref{bGlemma}.
\begin{corollary}\label{bGlemma}
The optimal classifier $C$ will have $p_C(y=K+1|x\sim p_u(x)) = 0$ and $p_C(y=K+1|x\sim p_{bG}(x)) = 1$. 
\end{corollary}

Corollary \ref{bGlemma} indicates that optimal $C$ will put $bG$ generated images into $K + 1$ class (\ie, ``fake'' class), while put unlabeled data into real classes. 

\vspace{-1mm}\subsubsection{Non-increasing Divergence Property} \label{EM_framework}\vspace{-1mm}
Our goal is to estimate the conditional distribution $p(y|x)$ with a parameterized $C$ modeled as $p_{\theta}(y|x,y\leq K)$. The objective function can be written as minimizing $\text{KL}(p(y|x)||p_{\theta}(y|x,y\leq K))$. In the SSL setting, we only have part of the labels $y$, so we can thus rewrite the problem as minimizing $\text{KL}(p(y_l|x)||p_{\theta}(y_l|x,y\leq K))$.
One natural way to facilitate the convergence rate is using the EM algorithm to first infer the label of $x_u$ and then update based on the complete data \cite{nigam2006semi}. In our four-player game, in addition to the predicted label $y_u$ from unlabelled data $x_u$, we further introduce $(x_{gG},y_{gG})$ pairs from $gG$ as latent variables, denoted as $Z = \{x_{gG},y_{gG},y_u\}$. We then interpret our mechanism from a variational view of the EM algorithm to illustrate the non-increasing property of the KL divergence. 

\emph{\textbf{Property I.}} Chain rule of KL divergence:
\begin{equation}
    \text{KL}(P(X,Z)||P_{\theta}(X,Z)) = \text{KL}(P(X)||P_{\theta}(X))+\mathbb{E}_{x\sim P(X)}[\text{KL}(P(Z|x)||P_{\theta}(Z|x))].
\end{equation}
By \emph{\textbf{Property I}}, we can rewrite our objective function as:
\begin{equation}
    \min_\theta\text{KL}(p(y_l|x)||p_\theta(y_l|x,y\leq K)) = \min_\theta\min_{p(Z|x)}\text{KL}(p(y_l,Z|x)||p_\theta(y_l,Z|x,y\leq K)),
\end{equation}
which is an iterative minimization procedure. Following the EM algorithm, we have an \textit{E-step} and an \textit{M-step} in UGAN. More specifically, for the \textit{E-step} at the $s$th iteration, given parameters $\theta_s$ of $C$, we have:
\begin{equation}\label{Estep}
    p(Z|x) = p_{\theta_s}(Z|x) = p_{gG}(x_{gG},y_{gG}|x_u,x_l,y_u,y_l)p_{\theta_s}(y_u|x_u),
\end{equation}
which indicates the procedure that $C$ first predicts labels for unlabelled data, and then sends them to $D$ and $gG$ to generate good pseudo pairs $(x_{gG},y_{gG})$. After gathering the latent variables, the \textit{M-step} is: 
\begin{equation}\label{Mstep}
\begin{aligned}
    \theta_{s+1} &= \text{argmin}_{\theta}\text{KL}(p(y_l,Z|x)||p_\theta(y_l,Z|x,y\leq K))\\&=\text{argmax}_\theta\mathbb{E}_{(y_l,Z|x)\sim p_{\theta_s} (Z|x_u)p_l(y_l|x_l)}[\log p_\theta(y_l,Z|x,y\leq K)],
\end{aligned}
\end{equation}
which will result in $\theta_{s+1}$ being the MLE based on the data at current iteration $s$. 

By applying the EM mechanism, we can inherit its non-increasing property which is stated in the following Corollary \ref{EMproperty}.
\begin{corollary}\label{EMproperty}
If applying the iterative procedure described in (\ref{Estep}) and (\ref{Mstep}), and the exact maximization can be obtained at (\ref{Mstep}) for each iteration, then
\begin{equation}
\text{KL}(p(y_l|x)||p_{\theta_{s+1}}(y_l|x,y\leq K))\leq \text{KL}(p(y_l|x)||p_{\theta_s}(y_l|x,y\leq K))
\end{equation}
\end{corollary}

\vspace{-3mm}\section{Experiments and Discussion}\vspace{-3mm} \label{exp_and_discuss}
We now present UGAN's performance on MNIST \cite{lecun1998gradient}, SVHN \cite{netzer2011reading}, and CIFAR10 \cite{krizhevsky2009learning} datasets (see details of datasets in Appendix \ref{app_datasets}). We implement our model based on Tensorflow 1.10 \cite{girija2016tensorflow} and optimize it on NVIDIA Titan X GPUs. The detailed architecture can be found in Appendix \ref{app_architecture}. The $gG$ generated images is not applied until the number of epochs reaches a threshold such that $gG$ can generate reliable image-lable pairs. For MNIST and SVHN, we choose 200, while for CIFAR10 we choose 400. Batch size is an important parameter that affects model performance \cite{li2019semi}. In our experiments, we use 50 for $bG$ on MNIST and SVHN, 25 for $bG$ on CIFAR10. For $gG$, we fix batch size as 100. All of the other hyperparameters including relative weights and parameters in Adam \cite{kingma2014adam} are fixed according to \cite{salimans2016improved,chongxuan2017triple,dai2017good} across all of the experiments.

\vspace{-3mm}\begin{figure*}[!hb]
\begin{center}
  \includegraphics[width=13.6cm]{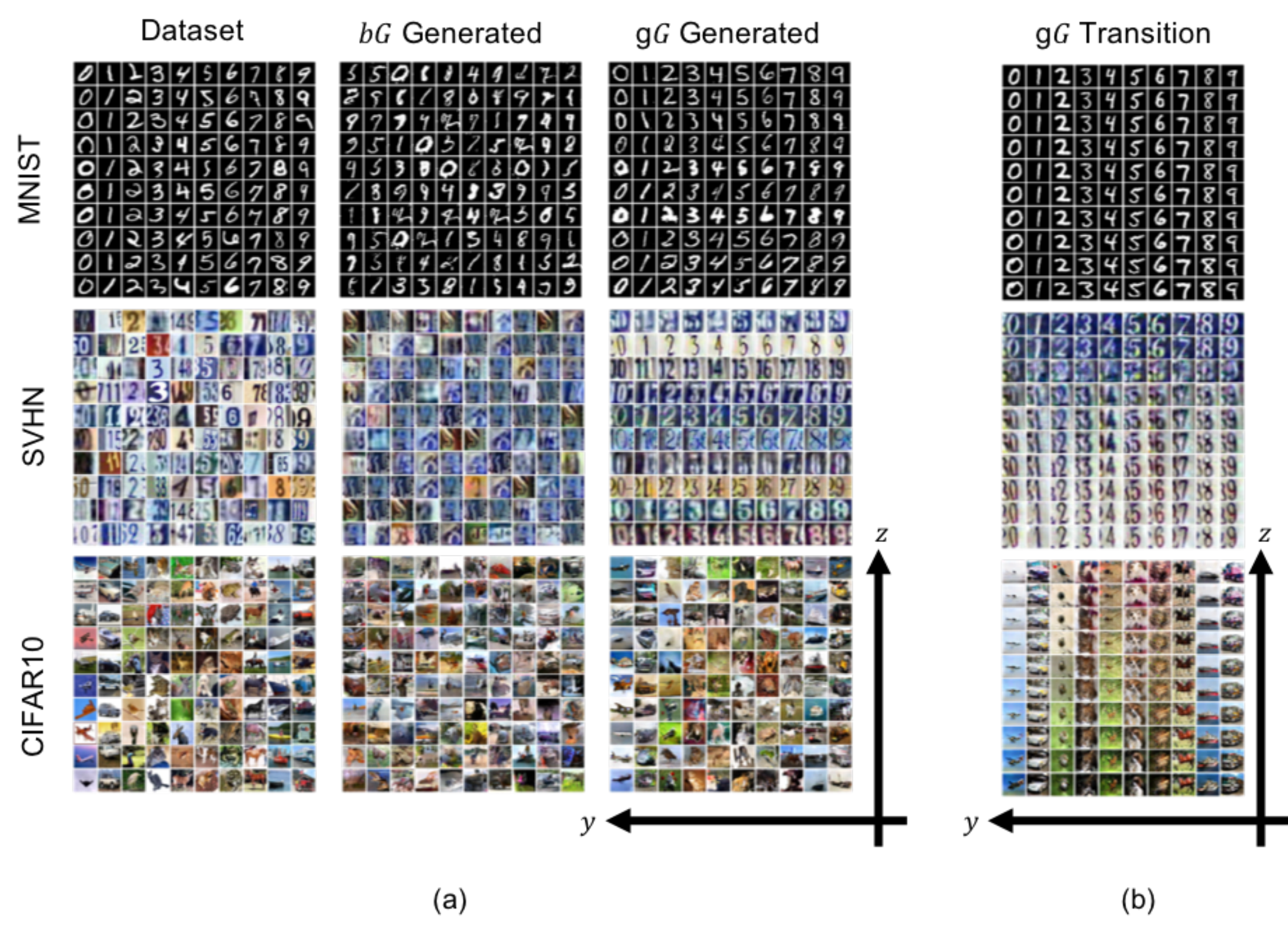}\vspace{-3mm}
  \caption{(a) Left: randomly selected data from datasets; mid: $bG$ generated images; right: $gG$ generated images sampled by varying the class label $y$ in the horizontal axis and the latent vectors $z$ in the vertical
    axis. (b) Class-conditional latent space interpolation. The vertical axis is the direction for latent vector interpolation, while the horizontal axis for varying the class labels.} \label{Figure2} \vspace{-5mm}
\end{center}
\end{figure*}

\vspace{-0mm}\subsection{Classification} \label{classification}\vspace{-2mm}
We report our classification accuracy, along with state-of-the-art methods on benchmark datasets in Table \ref{Table1}. Our results show that UGAN consistently improves performance, and achieves state-of-the-art results on all of the datasets without the use of data augmentation, such as rotation, flip, \etc. 

\vspace{-2mm}\begin{table}[!htbp]
\caption{Comparison with state-of-the-art methods on three benchmark datasets. Only methods without data augmentation are included. Results are averaged over 10 runs. }
\label{Table1}
\centering
\begin{tabular}{lccc}
\hline
Methods    & MNIST $n = 100$        & SVHN $n = 1000$         & CIFAR10 $n = 4000$      \\ \hline \hline
CatGAN\cite{springenberg2015unsupervised}     & $98.09 \pm 0.1\%$  & -            & $80.42 \pm 0.46\%$ \\
ALI \cite{dumoulin2016adversarially}        & -            &$ 92.58 \pm 0.65\%$ & $82.01 \pm 1.62$   \\
VAT \cite{miyato2018virtual}       & $98.64\%$      & $93.17\%$      & $85.13\%$      \\
$\Pi$ Model \cite{laine2016temporal}   & -            & $94.57 \pm 0.25\%$ & $83.45 \pm 0.29\%$ \\ \hline
FM-GAN \cite{salimans2016improved}    & $99.07 \pm 0.07\%$ & $91.89 \pm 1.3\%$  & $81.37 \pm 2.32\%$ \\
Triple-GAN \cite{chongxuan2017triple} & $99.09 \pm 0.58\%$ & $94.23 \pm 0.17\%$ & $83.01 \pm 0.36\%$ \\
Bad-GAN \cite{dai2017good}   & $99.21 \pm 0.10\%$ &$ 95.75 \pm 0.03\%$ & $85.59 \pm 0.30\%$ \\ \hline
UGAN  &$\mathbf{99.21 \pm 0.08\%}$    & $\mathbf{96.49 \pm 0.09\%}$ & $\mathbf{85.66 \pm 0.06\%}$ \\ \hline
\end{tabular}
\end{table}\vspace{-2mm}

\vspace{-2mm}
\begin{table*}[!htbp]
\caption{Test accuracy on semi-supervised MNIST. Results are averaged over 10 runs. $*$ denotes hand selection of labeled data. $\dag$ denotes our implementation of the model.}
\label{Table2}
\centering
\begin{tabular}{lcccc}
\hline
Model & \multicolumn{4}{c}{\begin{tabular}[c]{@{}c@{}}Test accuracy for a given number of labeled samples\end{tabular}} \\& 20& 50& 100& 200\\ \hline \hline
FM-GAN \cite{salimans2016improved}                & $83.23\pm4.52\%$ &  $97.79\pm1.36\%$                           & $99.07\pm0.07\%$      & $99.10\pm0.04\%$                            \\
Bad GAN \cite{dai2017good}                &-                            &-                            & $99.21\pm0.10\%$      &-                            \\
Triple-GAN \cite{chongxuan2017triple}            & $95.19\pm4.95\%$                & $98.44\pm0.72\%$               & $99.09\pm0.58\%$               & $99.33\pm0.16\%$      \\ \hline
$\text{Bad GAN}^{\dag}$         & $88.38\pm3.08\% ^*$                     & $96.24\pm0.16\%$                    & $99.17\pm0.03\%$                    & $99.20\pm0.03\%$                    \\
$\text{Triple-GAN}^{\dag}$        & $95.93\pm4.45\%^{*}$       & $98.68\pm1.12\%$      & $99.07\pm0.46\%$               & $99.17\pm0.08\%$               \\ \hline
UGAN       & $\mathbf{97.34\pm6.86\%^{*}}$       & $\mathbf{98.92\pm0.13\%}$      & $\mathbf{99.21\pm0.08\%}$               & $\mathbf{99.35\pm0.05}\%$              \\ \hline
\end{tabular}
\end{table*}\vspace{-2mm}
To further understand our model's behavior over different numbers of labeled data, we re-implemented Triple-GAN and Bad GAN, and performed an extensive investigation by varying the amount of labeled data. Following common practice, this was done by omitting different amounts of the underlying labeled dataset \cite{salimans2016improved, pu2016variational, sajjadi2016mutual, tarvainen2017mean}. The labeled data used for training were randomly selected stratified samples unless otherwise specified. For fair comparison, we used  the  same  network  architecture for each component in all models (see Appendix \ref{app_architecture}). Table \ref{Table2} shows the results of the experiments on MNIST. The similarity of our results to those reported in the original papers suggests that our reproduced models are accurate instantiations of Triple-GAN and Bad GAN. We observe that with a medium amount of labeled data (\textit{e.g.}, MNIST $n=100$), Bad GAN performs better than Triple-GAN. However, with smaller amounts of labeled data, Triple-GAN performs better, which demonstrates that it is less sensitive to the amount of labeled data than Bad GAN. UGAN inherits the good properties from both of them, resulting in a constant improvement across all cases (see results on SVHN and CIFAR10 in Appendix \ref{app_label_amount}). Another interesting observation is that the selection of labeled data plays a crucial role in the low-labeled data regime, that is, selecting representative labeled data with which to train is the key to achieving good performance. This issue is further discussed in Appendix \ref{app_label_importance}.

\vspace{-2mm}\subsection{Image Generation}\vspace{-2mm}
UGAN is able to train a $gG$ and a $bG$ simultaneously (see an evolution of the generated images in Appendix \ref{app_evolution}). In Fig. \ref{Figure2} (a), we show the images generated by $gG$ and $bG$ after training. Our $gG$ is able to generate clear images and meaningful samples conditioned on class labels, while $bG$ generates ``bad'' images that look like a fusion of samples from different classes. We quantitatively evaluate generated samples on CIFAR10 via the inception score following Salimans \etal. \cite{salimans2016improved}. The value of $gG$ generated samples is $4.19 \pm 0.07$, while that of $bG$ generated samples is $3.31 \pm 0.02$. In addition, $gG$ retains Triple-GAN's advantage in that it is able to disentangle classes and styles. In Fig. \ref{Figure2}(a), the $gG$ generated images are sampled by varying the class label $y$ in the horizontal axis and the latent vectors $z$ in the vertical axis. The latent vector $z$ encodes meaningful physical appearances, such as scale, intensity, orientation, color, \etc, while the label $y$ controls the semantics of the generated images. Furthermore, $gG$ can transition smoothly from one style to another with different visual factors without losing the label information as shown in Fig. \ref{Figure2} (b). This demonstrates that $gG$ can learn meaningful latent representations instead of simply memorizing the training data.

\vspace{-2mm}\subsection{Effectiveness of Good and Bad Generators}\vspace{-2mm} \label{gGbGeffectiveness}
\begin{figure*}
\begin{center}
  \includegraphics[width=13.5cm]{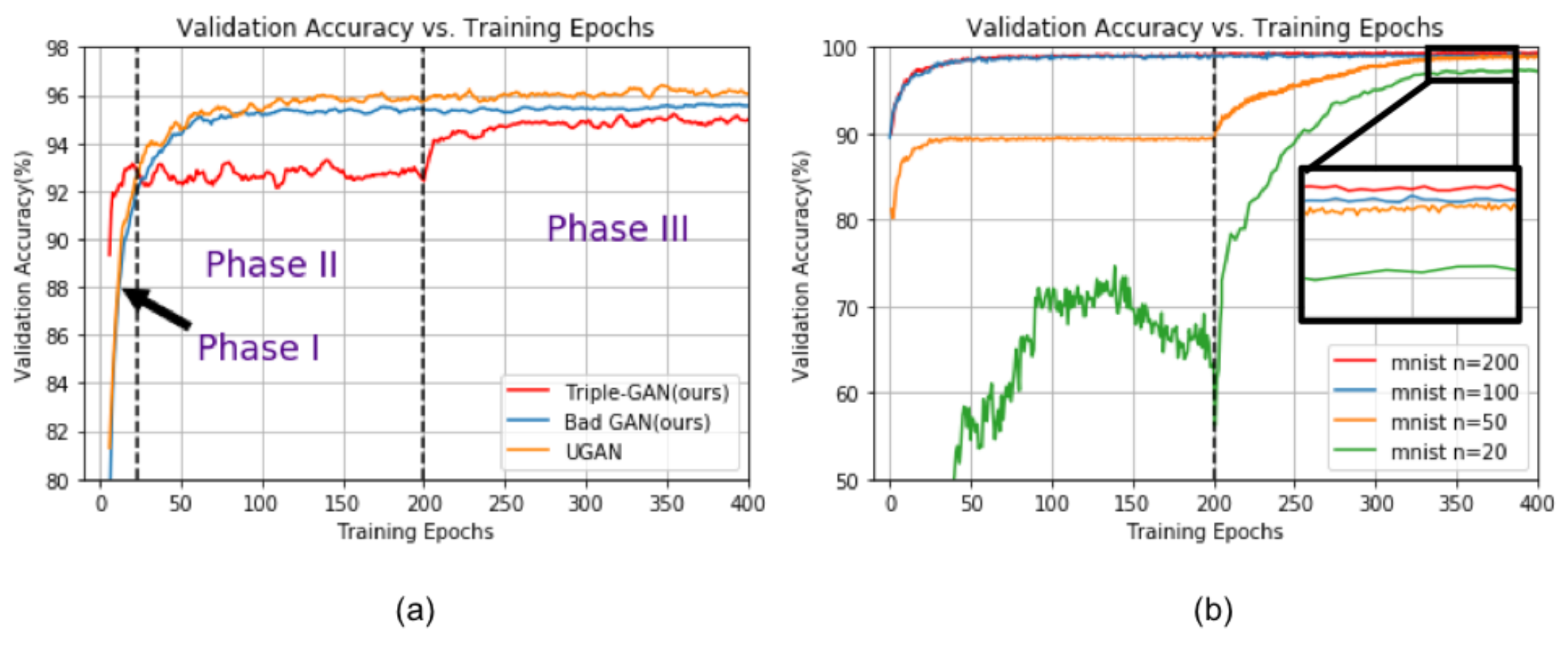}\vspace{-3mm}
  \caption{(a) Comparison of Validation Accuracy vs. Training Epochs on our implemented Triple-GAN, Bad GAN, and UGAN. The experiments are performed on SVHN $n = 1000$. (b) UGAN Validation Accuracy vs. Training Epochs under various amounts of labeled data on MNIST.} \label{Figure3}\vspace{-7mm}
\end{center}
\end{figure*}

As discussed in Section \ref{classification}, UGAN achieves consistent improvement across all the cases due to inheriting the best properties of Triple-GAN and Bad GAN. In Fig. \ref{Figure3} (a), we demonstrate a comparison of Validation Accuracy vs. Training Epochs for our implemented Triple-GAN, Bad GAN, and UGAN on SVHN $n=1000$. Note that for Triple-GAN, we trained it to 1000 epochs, but only show the first 400 epoch in the figure. Qualitatively, we observe three separate training phases:
\vspace{-2mm}
\begin{enumerate}[leftmargin=.2in, label=(\roman*)]
\vspace{-0.5mm}\item In \textbf{Phase I}, the performance of Bad GAN and UGAN are worse than Triple-GAN. We speculate this is due to the fact that Triple-GAN $C$ deals with a classification of $K$ classes, while Bad-GAN and UGAN, $C$ deal with $K + 1$ classes.
\vspace{-0.5mm}\item In \textbf{Phase II}, Bad GAN and UGAN start to surpass Triple-GAN, which indicates $bG$ generated samples start to exert an effect on the classification boundary. UGAN also performs better than Bad GAN in this phase thanks to the adversarial game that requires $C$ to produce reliable pseudo labels for unlabeled data to fool $D$. 
\vspace{-0.5mm}\item In \textbf{Phase III}, we start to use $gG$ generated samples to train $C$. UGAN surpasses both Triple-GAN and Bad GAN by a clear margin. From the perspective of $C$, $gG$ generates samples that are used to complement the lack of training data in SSL, $bG$ generated samples are used to force the decision boundary to lie in the correct place, and $D$ requires $C$ to keep moving itself toward the true data distribution $p(x)p_{C}(y|x, y \leq K) \approx p(x, y)$. All of these factors contribute to the final performance of UGAN.
\end{enumerate}\vspace{-2mm}

Similar observations can also be found in Appendix \ref{app_generator_effect} on MNIST and CIFAR10. Moreover, we hypothesize that for fewer labeled data, $gG$ plays an important role, as $gG$ is able to model the class-aware data distribution under weak supervision and use them to complement the lack of the training samples. While for larger labeled data, $bG$ plays a more important role by generating complementary samples and forcing the decision boundary to lie between the data manifolds of different classes. Empirically, we show our model's validation accuracy under various amounts of labeled data on MNIST in Fig. \ref{Figure3} (b). As can be seen, when we push the number of labeled data to extremely low numbers, the training curve becomes more like that in Triple-GAN \ie, a bump is shown clearly at $\text{epoch}=200$ when we start to use $gG$ generated samples to train $C$. However, we do not find a similar transition on SVHN and CIFAR10 (see Appendix \ref{app_generator_effect}). One possible explanation is that when we use too few labeled data, $gG$ fails to model the conditional distribution due to the complexity of SVHN and CIFAR10. Note that we only used traditional techniques for training the GAN. With recent advances in generating high quality images using GANs \cite{brock2018large, miyato2018spectral, lucic2019high}, our model may be able to achieve further performance improvements on more complex datasets with even fewer labeled data.

\vspace{-3mm}\section{Conclusions}\vspace{-3mm}
We have presented unified-GAN (UGAN), a new GAN framework 
for semi-supervised learning. By learning from good and bad samples through adversarial training, we have demonstrated that our model performs better on image classification tasks across several benchmark datasets and under a range of labeled training data. We envision that UGAN can be used in a variety of scenarios, such as healthcare, where obtaining labeled data can be expensive and time-consuming. 



{
\bibliographystyle{abbrv}
\bibliography{neurips_2019.bib}
}
\newpage
\appendixpage
\appendix
\section{Loss Function of the Classifier}\label{app_softmax}
Softmax layer is over-parameterized, therefore we can still model $C$ with $K$ neurons at the output layer. To represent $K+1$ classes, the loss function should be modified as detailed below.

First let us rewrite the four components of $C$'s objective function:
\begin{equation} \label{classifier_loss_sep_app1}
\begin{aligned}
L_{C_{1}} &= -\mathbb{E}_{x, y\sim p_{l}(x, y)}[\log (p_{C}(y|x, y \leq K)]
& L_{C_2} &=-\mathbb{E}_{x, y\sim p_{gG}(x, y)}[\log (p_{C}(y|x, y \leq K)]\\
L_{C_{3}} &= -\mathbb{E}_{x \sim p_{u}(x)}[\log (1 - p_{C}(y = K + 1 | x)]
& L_{C_{4}} &= -\mathbb{E}_{x \sim p_{bG}(x)}[\log ( p_{C}(y = K + 1 | x)]\\
\end{aligned}
\end{equation}

Suppose $\{l_1(x), l_2(x), l_3(x), \cdot \cdot \cdot, l_K(x), l_{K + 1}(x)\}$ represents the logits before the softmax-layer for input $x$, by using the fact that softmax is over-parameterized, we can fix the logit $l_{K+1}(x) = 0 ~ \forall x$ for the $bG$ generated images and the output of the softmax remains the same. Hence, we can reformulate the above four components as
\begin{equation} \label{classifier_loss_sep_app2}
\begin{aligned}
L_{C_{1}} &= -\mathbb{E}_{x, y\sim p_{l}(x, y)}[-l_y + \log (\sum_{i = 1}^K \text{exp}l_i)]\\
L_{C_2} &=-\mathbb{E}_{x, y\sim p_{gG}(x, y)}[-l_y + \log (\sum_{i = 1}^K \text{exp}l_i)]\\
L_{C_{3}} &= -\mathbb{E}_{x \sim p_{u}(x)}[-\log(\sum_{i=1}^K \text{exp} l_i) + \log(1 + \sum_{i = 1}^{K}\text{exp} l_i)]\\
L_{C_{4}} &= -\mathbb{E}_{x \sim p_{bG}(x)}[\log (1 + \sum_{i=1}^K \text{exp} l_i)]\\
\end{aligned}
\end{equation}
Define the log sum exponent function as $\text{LSE}(\bm{x}) = \log (\sum_{j}\exp x_j)$ and softplus function as $\text{softplus}(x) = \log (1 + \exp x)$, the losses can be further simplified as
\begin{equation} \label{classifier_loss_sep_app3}
\begin{aligned}
L_{C_{1}} &= -\mathbb{E}_{x, y\sim p_{l}(x, y)}[-l_y + \text{LSE}(\bm{l})]\\
L_{C_2} &=-\mathbb{E}_{x, y\sim p_{gG}(x, y)}[-l_y + \text{LSE}(\bm{l})]\\
L_{C_{3}} &= -\mathbb{E}_{x \sim p_{u}(x)}[-\text{LSE}(\bm{l}) + \text{softplus}(\text{LSE}(\bm{l}))]\\
L_{C_{4}} &= -\mathbb{E}_{x \sim p_{bG}(x)}[\text{softplus}(\text{LSE}(\bm{l}))]\\
\end{aligned}
\end{equation}
which are used in our code implementation.
\section{Detailed Theoretical Analysis} \label{app_theoretical_analysis}
\textbf{Lemma 3.1}
\textit{For any fixed $C$ and $G$, the optimal $D$ of the game defined by the loss function (\ref{good_generator_loss})-(\ref{classifier_loss}) is}
\begin{equation}
    D^*_{C,gG,bG}(x,y) = \frac{p_l(x,y)}{p_l(x,y)+p_{\frac{1}{2}}(x,y)},
\end{equation}
\textit{where $p_{\frac{1}{2}}(x,y) = \frac{1}{2}p_{gG}(x,y) + \frac{1}{2}p_{C}(x,y|y\leq K)$.}

\textit{Proof}: This follows from Proposition 1 of \cite{goodfellow2014generative} directly. 

\textbf{Theorem 3.2}
\textit{The global minimum of $V(C, gG, bG)$ is achieved only when $p_l(x,y)=p_{gG}(x,y)=p_C(x,y|y\leq K)$.}

\textit{Proof}:

Given $D^*_{C,gG,bG}$, we can reformulate our value function as
\begin{equation} 
V(C,gG,bG) = -\log 4+2JSD(p_l(x,y),
p_{\frac{1}{2}}(x,y))+L_{C1}+L_{C2}+L_{C3}+L_{C4}.
\end{equation}
We first focus on the term with respect to $p_C(x,y|y\leq K)$, denoted the corresponding loss as $\Tilde{V}(C|y\leq K)$, we have
\begin{equation}
\begin{aligned}
    \Tilde{V}(C|y\leq K) &\propto 2JSD(p_l(x,y),
p_{\frac{1}{2}}(x,y)) -\mathbb{E}_{x, y\sim p_{l}(x, y)}[\log (p_{C}(y|x, y \leq K)]\\&-\mathbb{E}_{x, y\sim p_{gG}(x, y)}[\log (p_{C}(y|x, y \leq K)]\\
&\propto 2JSD(p_l(x,y),p_{\frac{1}{2}}(x,y))+KL(p_\beta (x,y)|| p_{C}(y|x, y \leq K),
\end{aligned}
\end{equation}
where $p_\beta(x,y) = \beta p_l(x,y)+(1-\beta)p_{gG}(x,y)$ and $\beta/(1-\beta)$ is the ratio of data we feed into classifier between true labeled data and data pairs from good generator. Therefore the global minimum can only be achieved when 
\begin{equation}
    \begin{aligned}
    p_l(x,y) = \frac{1}{2}p_{gG}(x,y) + \frac{1}{2}p_{C}(x,y|y\leq K)\\
    p_{C}(x,y|y\leq K) = \beta p_l(x,y)+(1-\beta)p_{gG}(x,y),
    \end{aligned}
\end{equation}
and it is obtained when $p_l(x,y)=p_{gG}(x,y)=p_C(x,y|y\leq K)$.

\textbf{Corollary 3.2.1}
\textit{The optimal classifier $C$ will have $p_C(y=K+1|x\sim p_u(x)) = 0$ and $p_C(y=K+1|x\sim p_{bG}(x)) = 1$.}

\textit{Proof}: Because $p_C(y=K+1|x)$ and $p_C(y|x,y\leq K)$ are independent, we can consider them separately. The term related to $p_C(y=K+1|x)$ in loss function is 
\begin{equation}
    L_{C_3}+L_{C_4} = -\mathbb{E}_{x \sim p_{u}(x)}[\log (1 - p_{C}(y = K + 1 | x)] -\mathbb{E}_{x \sim p_{bG}(x)}[\log ( p_{C}(y = K + 1 | x)],
\end{equation}
which achieves its minimal $0$ when $p_C(y=K+1|x\sim p_u(x)) = 0$ and $p_C(y=K+1|x\sim p_{bG}(x)) = 1$.

\textbf{Corollary 3.2.2}
\textit{If applying the iterative procedure described in (\ref{Estep}) and (\ref{Mstep}),}
\begin{equation}
\text{KL}(p(y_l|x)||p_{\theta_{s+1}}(y_l|x,y\leq K))\leq \text{KL}(p(y_l|x)||p_{\theta_s}(y_l|x,y\leq K))
\end{equation}

\textit{Proof:} Define 
\begin{equation}
    J(\theta, p(Z|x)) = \text{KL}(p(y_l|x)p(Z|x)||p_\theta(y_l,Z|x,y\leq K)),
\end{equation}
and 
\begin{equation}
    J(\theta) = \text{KL}(p(y_l|x)||p_\theta(y_l|x,y\leq K)).
\end{equation}
Then we have
\begin{equation}
    J(\theta_{s+1})\leq J(\theta_{s+1},p_{\theta_s}(Z|x))\leq J(\theta_s, p_{\theta_s}(Z|x)) = J(\theta_s).
\end{equation}

\section{Datasets}\label{app_datasets}
We apply UGAN on the widely adopted MNIST \cite{lecun1998gradient}, SVHN \cite{netzer2011reading}, and CIFAR10 \cite{krizhevsky2009learning} datasets. MNIST consists of 50,000 training samples, 10,000 validation samples, and 10,000 testing samples of handwritten digits of size $28 \times 28$. SVHN consists of 73,257 training samples and 26,032 testing samples. Each sample is a colored image of size $32 \times 32$, containing a sequence of digits with various backgrounds. CIFAR10 consists of colored images distributed across 10 general classes -- \textit{airplane}, \textit{automobile}, \textit{bird}, \textit{cat}, \textit{deer}, \textit{dog}, \textit{frog}, \textit{horse}, \textit{ship} and \textit{truck}. It contains 50,000 training samples and 10,000 testing samples of size $32 \times 32$. Following \cite{chongxuan2017triple}, we reserve 5,000 training samples from SVHN and CIFAR10 for validation if needed in our experiments. 

\section{Network Architecture}\label{app_architecture}
We list the detailed architecture we used to construct UGAN in Table \ref{Table1app}, Table \ref{Table2app} and Table \ref{Table3app} respectively. To re-implement Triple-GAN and Bad GAN, we also use the same architecture of the corresponding parts for fair comparison. Note that in Bad GAN, the discriminator has two roles: to classify the real data into the right class and to distinguish the real samples from the fake samples.  For clarity, we refer to Bad GAN's $D$ as $C$ in the table, while $D$ is a conditional network that presents in Triple-GAN and UGAN.

\begin{table}[!ht]
\caption{MNIST}
\label{Table1app}
\centering
\begin{tabular}{c|c|c|c}
\hline
$bG$        & $gG$        & $C$ & $D$ \\ \hline
$z \sim p(z)$ & $y \sim p(y)$, $z \sim p(z)$ & $x \sim p_{\{l, u, gG, bG\}}(x)$& $(x, y) \sim p_{\{l,gG,C\}}(x, y)$\\ \hline
\multicolumn{2}{c|}{\begin{tabular}[c]{@{}c@{}}MLP 500 units, \\ softplus, batch norm\\ \\ MLP 500 units, \\softplus, batch norm \\ \\ MLP 500 units, \\ softplus, batch norm\end{tabular}} & \begin{tabular}[c]{@{}c@{}}MLP 1000 units, lRelu, \\ Gaussian noise, weight norm\\ MLP 500 units, lRelu, \\ Gaussian noise, weight norm\\ MLP 250 units, lRelu, \\ Gaussian noise, weight norm\\ MLP 250 units, lRelu, \\ Gaussian noise, weight norm\\ MLP 250 units, lRelu, \\ Gaussian noise, weight norm\\ MLP 10 units, softmax, \\ Gaussian noise, weight norm\end{tabular} & \begin{tabular}[c]{@{}c@{}}MLP 1000 units, lRelu, \\ Gaussian noise, weight norm\\ MLP 500 units, lRelu, \\ Gaussian noise, weight norm\\ MLP 250 units, lRelu, \\ Gaussian noise, weight norm\\ MLP 250 units, lRelu, \\ Gaussian noise, weight norm\\ MLP 250 units, lRelu, \\ Gaussian noise, weight norm\\ MLP 12 units, sigmoid, \\ Gaussian noise, weight norm \end{tabular}  
\\ \hline
\end{tabular}
\end{table}

\begin{table}[!ht]
\caption{SVHN}
\label{Table2app}
\centering
\begin{tabular}{c|c|c|c}
\hline
$bG$        & $gG$        & $C$ & $D$ \\ \hline
$z \sim p(z)$ & $y \sim p(y)$, $z \sim p(z)$ & $x \sim p_{\{l, u, gG, bG\}}(x)$& $(x, y) \sim p_{\{l,gG,C\}}(x, y)$\\ \hline
\multicolumn{2}{c|}{\begin{tabular}[c]{@{}c@{}}MLP 8192 units, \\ Relu, batch norm\\ Reshape $512 \times 4 \times 4$\\ \\ $5 \times 5$ deconv. 256. stride 2, \\ Relu, batch norm\end{tabular}} & \begin{tabular}[c]{@{}c@{}}Gaussian noise, 0.2 dropout\\ $3 \times 3$ conv. 64. \\lRelu, weight norm\\$3 \times 3$ conv. 64. \\lRelu, weight norm\\$3 \times 3$ conv. 64. lRelu,  \\ stride 2, weight norm\\ 0.5 dropout\end{tabular} & \begin{tabular}[c]{@{}c@{}}0.2 dropout\\ $3 \times 3$ conv. 32. \\lRelu, weight norm\\ $3 \times 3$ conv. 32. lRelu, \\stride 2, weight norm\\ 0.2 dropout\end{tabular}
\\ \hline
\multicolumn{2}{c|}{\begin{tabular}[c]{@{}c@{}}$5 \times 5$ deconv. 128. stride 2, \\ Relu, batch norm\end{tabular}} & \begin{tabular}[c]{@{}c@{}}$3 \times 3$ conv. 128. \\lRelu, weight norm\\ $3 \times 3$ conv. 128. \\lRelu, weight norm\\ $3 \times 3$ conv. 128. lRelu, \\stride 2, weight norm\\ 0.5 dropout\end{tabular} & \begin{tabular}[c]{@{}c@{}}$3 \times 3$ conv. 64. \\lRelu, weight norm\\ $3 \times 3$ conv. 64. lRelu, \\stride 2, weight norm\\ 0.2 dropout\end{tabular}
\\ \hline
\multicolumn{2}{c|}{\begin{tabular}[c]{@{}c@{}}$5 \times 5$ deconv. 3. stride 2, \\ sigmoid, weight norm\end{tabular}} & \begin{tabular}[c]{@{}c@{}}$3 \times 3$ conv. 128. \\lRelu, weight norm\\ $3 \times 3$ conv. 128. \\lRelu, weight norm\\ $3 \times 3$ conv. 128. \\lRelu, weight norm\\ \\ Global pool\\ MLP 10 units,\\ softmax, weight norm\end{tabular} & \begin{tabular}[c]{@{}c@{}}$3 \times 3$ conv. 128. \\lRelu, weight norm\\ $3 \times 3$ conv. 128. \\lRelu, weight norm\\ \\ Global pool\\ MLP 1 unit, \\sigmoid, weight norm \end{tabular}
\\ \hline
\end{tabular}
\end{table}

\begin{table}[!ht]
\caption{CIFAR10}
\label{Table3app}
\centering
\begin{tabular}{c|c|c|c}
\hline
$bG$        & $gG$        & $C$ & $D$ \\ \hline
$z \sim p(z)$ & $y \sim p(y)$, $z \sim p(z)$ & $x \sim p_{\{l, u, gG, bG\}}(x)$& $(x, y) \sim p_{\{l,gG,C\}}(x, y)$\\ \hline
\multicolumn{2}{c|}{\begin{tabular}[c]{@{}c@{}}MLP 8192 units, \\ Relu, batch norm\\ Reshape $512 \times 4 \times 4$\\ \\ \\ $5 \times 5$ deconv. 256. stride 2, \\ Relu, batch norm\end{tabular}} & \begin{tabular}[c]{@{}c@{}}Gaussian noise, 0.2 dropout\\ $3 \times 3$ conv. 96. lRelu, weight norm\\$3 \times 3$ conv. 96. lRelu, weight norm\\$3 \times 3$ conv. 96. lRelu,  \\ stride 2, weight norm\\ 0.5 dropout\end{tabular} & \begin{tabular}[c]{@{}c@{}}0.2 dropout\\ $3 \times 3$ conv. 32. \\lRelu, weight norm\\ $3 \times 3$ conv. 32. lRelu, \\stride 2, weight norm\\ 0.2 dropout\end{tabular}
\\ \hline
\multicolumn{2}{c|}{\begin{tabular}[c]{@{}c@{}}$5 \times 5$ deconv. 192. stride 2, \\ Relu, batch norm\end{tabular}} & \begin{tabular}[c]{@{}c@{}}$3 \times 3$ conv. 192. \\lRelu, weight norm\\ $3 \times 3$ conv. 192. \\lRelu, weight norm\\ $3 \times 3$ conv. 192. lRelu, \\stride 2, weight norm\\ 0.5 dropout\end{tabular} & \begin{tabular}[c]{@{}c@{}}$3 \times 3$ conv. 64. \\lRelu, weight norm\\ $3 \times 3$ conv. 64. lRelu, \\stride 2, weight norm\\ 0.2 dropout\end{tabular}
\\ \hline
\multicolumn{2}{c|}{\begin{tabular}[c]{@{}c@{}}$5 \times 5$ deconv. 3. stride 2, \\ sigmoid, weight norm\end{tabular}} & \begin{tabular}[c]{@{}c@{}}$3 \times 3$ conv. 192. \\lRelu, weight norm\\ $3 \times 3$ conv. 192. \\lRelu, weight norm\\ $3 \times 3$ conv. 192. \\lRelu, weight norm\\ \\ Global pool\\ MLP 10 units,\\ softmax, weight norm\end{tabular} & \begin{tabular}[c]{@{}c@{}}$3 \times 3$ conv. 192. \\lRelu, weight norm\\ $3 \times 3$ conv. 192. \\lRelu, weight norm\\ \\ Global pool\\ MLP 1 unit, \\sigmoid, weight norm \end{tabular}
\\ \hline
\end{tabular}
\end{table}\vspace{-3mm}

\section{Results of Varying Amount of Labeled Data}\label{app_label_amount}
We perform our experiments on setups with 20, 50, 100, and 200 labeled examples in MNIST, 500, 1000, and 2000 labeled examples in SVHN, and 1000, 2000, 400, 8000 examples in CIFAR10. Table \ref{Table4app} $\sim$ \ref{Table5app} show the results of the experiemts on SVHN, and CIFAR10 respectively. We find that our UGAN constantly outperforms Triple-GAN and Bad GAN across a wide range of labled data.

\begin{table*}[!ht]
\caption{Test accuracy on semi-supervised SVHN. Results are averaged over 10 runs.}
\label{Table4app}
\centering
\begin{tabular}{cccc}
\hline
Model & \multicolumn{3}{c}{\begin{tabular}[c]{@{}c@{}}Test accuracy for \\ a given number of labeled samples\end{tabular}} \\& 500 & 1000 & 2000                         \\ \hline \hline
Bad GAN\cite{dai2017good}                & -                            & $95.75 \pm 0.03\%$             & -                   \\
Triple-GAN\cite{chongxuan2017triple}             & -                            & $94.23 \pm 0.17\%$                     & -                            \\ \hline
Bad GAN (ours)         & $94.21 \pm 0.45\%$                      & $95.32 \pm 0.07 \%$                                              & $95.47 \pm 0.39\%$             \\
Triple-GAN (ours)        & $94.67 \pm 0.12\%$             & $95.30 \pm 0.38\%$                                              & $95.37 \pm 0.09\%$                      \\\hline
UGAN       & $\mathbf{95.53 \pm 0.13\%}$             & $\mathbf{96.49 \pm 0.09\%}$                                              & $\mathbf{96.51 \pm 0.05\%}$                      \\ \hline
\end{tabular}
\end{table*}\vspace{-3mm}

\begin{table*}[!ht]
\caption{Test accuracy on semi-supervised CIFAR10. Results are averaged over 10 runs.}
\label{Table5app}
\centering
\begin{tabular}{ccccc}
\hline
Model & \multicolumn{4}{c}{\begin{tabular}[c]{@{}c@{}}Test accuracy for \\ a given number of labeled samples\end{tabular}} \\
                       & 1000                      & 2000                      & 4000                           & 8000                      \\ \hline \hline
Bad GAN \cite{dai2017good}                & -                         & -                         & $85.59\pm 0.03\%$                   & -                         \\
Triple-GAN \cite{chongxuan2017triple}             & -                         & -                         & $83.01 \pm 0.36\%$                   & -                         \\ \hline
Bad GAN (ours)         & $77.58 \pm 0.17\%$                   & $81.36 \pm 0.08\%$                   & $82.89 \pm 0.13\%$                        & $85.47 \pm 0.10\%$ \\ \hline         Triple-GAN (ours)          &      $81.08 \pm 0.57\%$              &   $81.79 \pm 0.37\%$    & $82.82 \pm 0.41\%$                        &    $85.37 \pm 0.18\%$                       \\ \hline
UGAN    &      $\mathbf{82.34 \pm 0.17\%}$              &   $\mathbf{83.88 \pm 0.13\%}$    & $\mathbf{85.66 \pm 0.06\%}$                        &    $\mathbf{86.58 \pm 0.09\%}$                       \\ \hline
\end{tabular}
\end{table*}

\section{Importance of Selected Labeled Data}\label{app_label_importance}
\begin{figure}
\begin{center}
  \includegraphics[width=13cm]{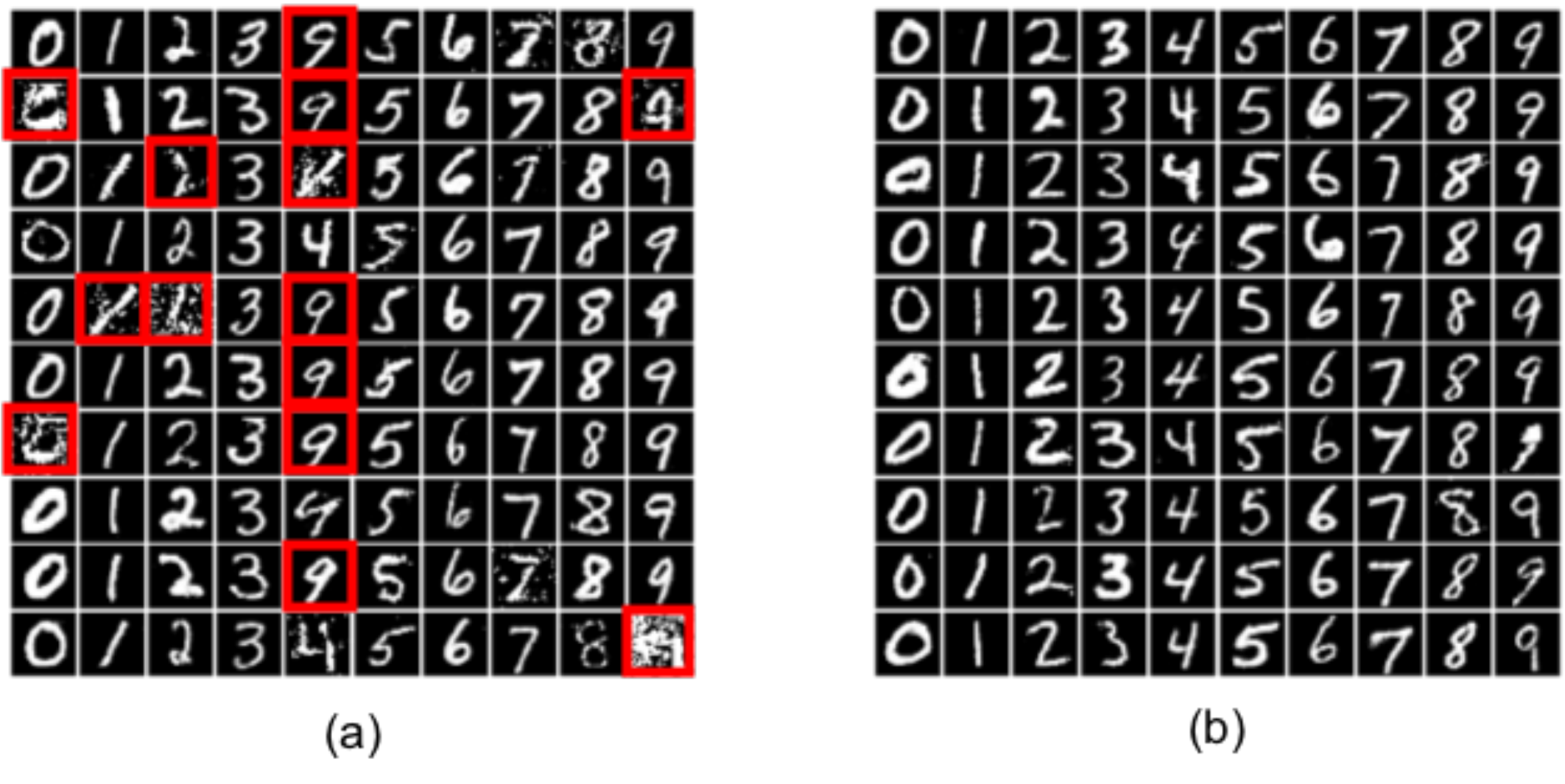}
  \caption{Two-runs of UGAN model on MNIST dataset. (a) A single run where we randomly select 20 labeled data. $gG$ generates a lot of wrong images conditioned on the label, resulting in bad performance of $C$. (b) Another run where we manually select 20 representative labeled examples. This time $gG$ is able to generate correct images, and $C$ achieves good classification performance.} \label{Figure4}
\end{center}
\end{figure}

One interesting observation is that the selection of labeled data plays a crucial role for training Triple-GAN, Bad GAN and UGAN in the low labeled data scenario. For most cases, the labeled data used for the training in our experiments are randomly selected stratified samples, except for the MNIST-20 case. In this case, we found selecting representative labeled data to train is the key to achieving good performance. The reported accuracy in Table \ref{Table2} is averaged over 10 runs where we manually selected different representative labeled data in a stratified way. Fig. \ref{Figure4}(a) shows a single run that UGAN uses randomly selected labeled data and does not achieve good results, while Fig. \ref{Figure4}(b) shows another run that is able to achieve higher accuracy. The failure of the first run is due to the initial selections for digit 4 being similar to 9, causing the generator to generate many 9s when conditioned on label 4. The generator also generates low-quality images. We also report that with a random selection of 20 labeled data, Tripe-GAN is able to achieve $76.78 \pm 6.47 \%$ accuracy over 3 runs, Bad GAN is achieving $68.12 \pm 0.60 \%$ over 10 runs, and UGAN is able to achieve $89.35 \pm 7.61\%$ accuracy over 3 runs. As can be seen, in both cases Triple-GAN outperfoms Bad GAN, while UGAN outperforms both of them, revealing that UGAN is least sensitive to the amounts of labeled data. The importance of selected labeled data is not surprising and is related to active learning, a potential future work could be extending UGAN for active learning.

\section{Generator Evolution}\label{app_evolution}
\begin{figure}
\begin{center}
  \includegraphics[width=13.8cm]{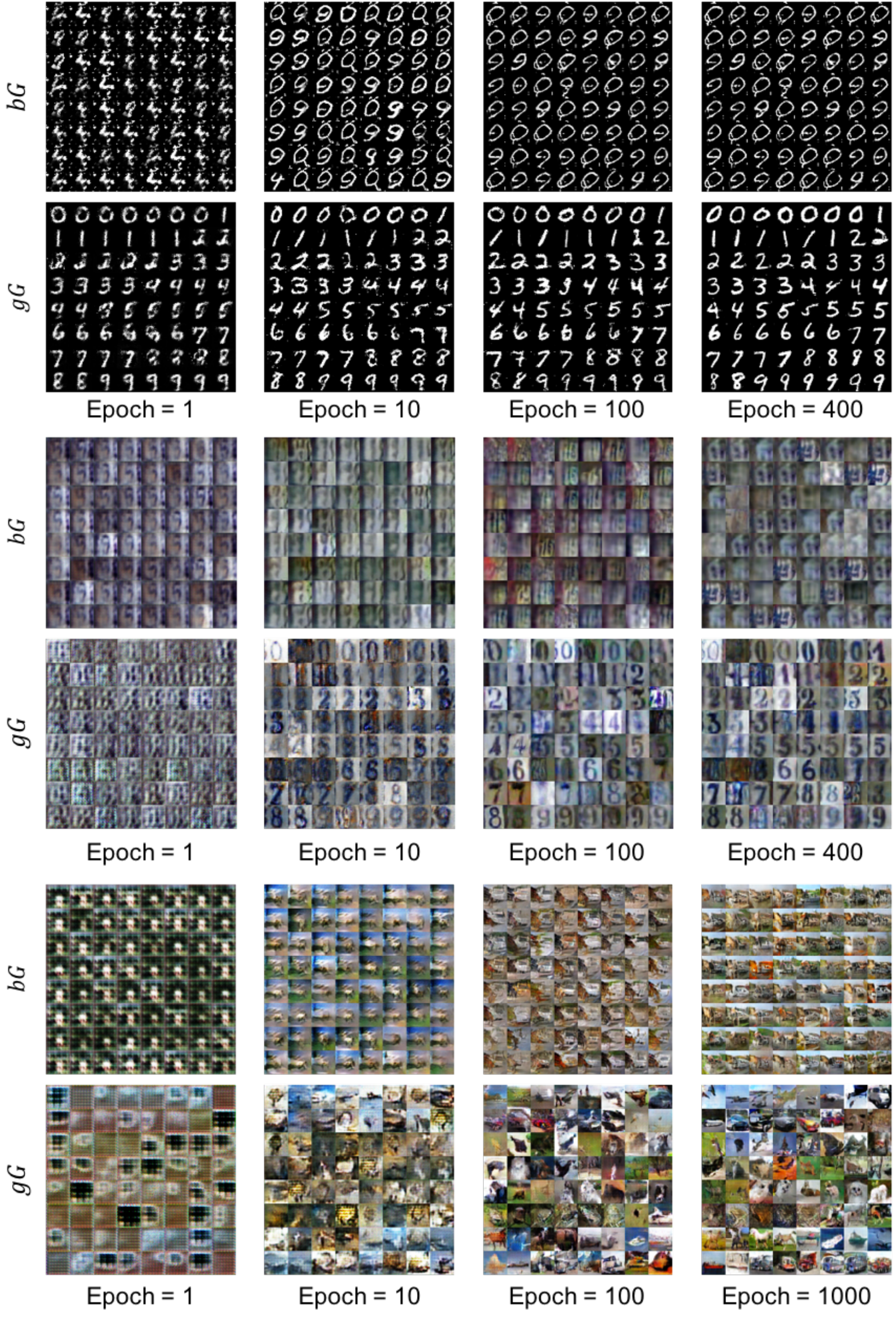}
  \caption{$gG$ and $bG$ evolution. Generated images from both $bG$ and $gG$ throughout training are shown. UGAN are trained on MNIST (upper), SVHN (middle), and CIFAR10 (lower). Through training, UGAN is able to obtain a good generator and a bad generator simultaneously.} \label{Figure5}
\end{center}
\end{figure}

By iteratively update $D$, $gG$, $C$, and $bG$ using gradient decent, UGAN is able to obtain a good generator and a bad generator simultaneously. To illustrate this, Fig.\ref{Figure5} shows an evolution of both $gG$ and $bG$ generated samples throughout the training on MNIST, SVHN, and CIFAR10. As the training progresses, $gG$ generated samples become clearer and semantic meaningful; $bG$ generated samples are more close to data manifold but semantic meaningless.

\section{Good and Bad Samples Effectiveness}\label{app_generator_effect}
\begin{figure}
\begin{center}
  \includegraphics[width=13.8cm]{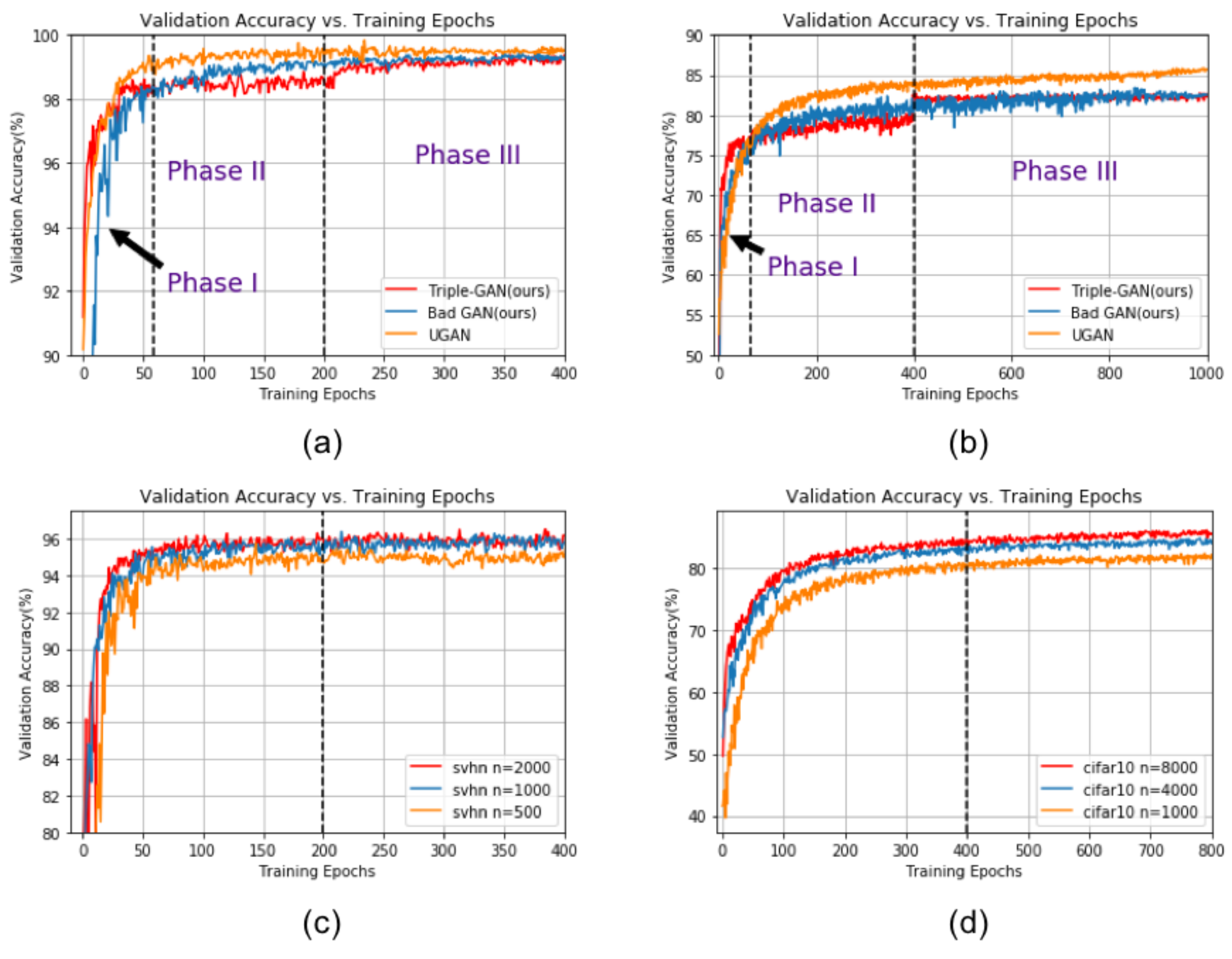}
  \caption{Comparison of Triple-GAN, Bad GAN, and UGAN on (a) MNIST $n = 100$ and (b) SVHN $n = 1000$. Similar three-phase training processes have been observed in both cases. UGAN Validation Accuracy vs. Training Epochs under various amount of labeled data on (c) SVHN and (d) CIFAR10. We don't find a similar transition on SVHN and CIFAR10 as in Fig. \ref{Figure3}(b). The vertical dot line in (c) and (d) denotes the epoch when we start to use $gG$ generated image-label pairs to train $C$.} \label{Figure6}
\end{center}
\end{figure}

As mentioned in Section \ref{gGbGeffectiveness}, we also observe a similar three phases training process in MNIST and CIFAR10. Fig. \ref{Figure6}(a) and (b) show the comparison among Triple-GAN, Bad GAN, and UGAN on MNIST and CIFAR10 respectively. The experiments are done under MNIST $n = 100$ and SVHN $n = 1000$.

For the number of labeled data effect, we don't find a similar transition on SVHN and CIFAR10 as in Fig. \ref{Figure3}(b). Instead, we find a graduate change of the learning curve under different amounts of labeled data. We also have tried to push the number of labeled data even low (\ie, $n < 500$ in SVHN and $n < 1000$ in CIFAR10), but UGAN fails to generate good image-label pairs. One possible explanation is that when we use too few labeled data, $gG$ fails to model the conditional distribution due to the complexity of SVHN and CIFAR10.

\end{document}